\documentclass[10pt, conference, compsocconf]{IEEEtran}
\usepackage{amsmath,epsfig}
\usepackage[mathscr]{euscript}
\usepackage{sidecap}
\usepackage[export]{adjustbox}
\usepackage{algorithm}
\usepackage[noend]{algpseudocode}
\usepackage{float}
\usepackage{caption}
\usepackage{graphicx}
\usepackage{subfig}
\usepackage{amsfonts}

\begin{document}\sloppy

\def\x{{\mathbf x}}
\def\L{{\cal L}}

\title{
Text2FaceGAN: Face Generation from Fine Grained Textual Descriptions}
%

\author{\IEEEauthorblockN{Osaid Rehman Nasir\textsuperscript{+}\IEEEauthorrefmark{1},
Shailesh Kumar Jha\textsuperscript{+}\IEEEauthorrefmark{1},
Manraj Singh Grover\IEEEauthorrefmark{1}, 
Yi Yu\IEEEauthorrefmark{2},
Ajit Kumar\IEEEauthorrefmark{3} and
Rajiv Ratn Shah\IEEEauthorrefmark{1}}
\IEEEauthorblockA{\IEEEauthorrefmark{1}MIDAS Lab, IIIT-Delhi\\
Delhi, India\\Email: midas@iiitd.ac.in}
\IEEEauthorblockA{\IEEEauthorrefmark{2}NII, Tokyo, Japan\\
Email: yiyu@nii.ac.jp}
\IEEEauthorblockA{\IEEEauthorrefmark{3}Adobe Systems\\Email: ajikumar@adobe.com}}



\maketitle

\begin{abstract}
Powerful generative adversarial networks (GAN) have been developed to automatically synthesize realistic images
from text. However, most existing tasks are limited to generating simple images such as flowers from captions. In this work, we extend this problem to the less addressed domain of face generation from fine-grained textual descriptions of face, \emph{e.g., ``A person has curly hair, oval face, and mustache"}. We are motivated by the potential of automated face generation to impact and assist critical tasks such as criminal face reconstruction. Since current datasets for the task are either very small or do not contain captions, we generate captions for images in the CelebA dataset by creating an algorithm to automatically convert a list of attributes to a set of captions. We then model the highly multi-modal problem of text to face generation as learning the conditional distribution of faces (conditioned on text) in same latent space. We utilize the current state-of-the-art GAN (DC-GAN with GAN-CLS loss) for learning conditional multi-modality. The presence of more fine-grained details and variable length of the captions makes the problem easier for a user but more difficult to handle compared to the other text-to-image tasks. We flipped the labels for real and fake images and added noise in discriminator. Generated images for diverse textual descriptions show promising results. In the end, we show how the widely used inceptions score is not a good metric to evaluate the performance of generative models used for synthesizing faces from text.
\end{abstract}
\section{Introduction}
\label{sec:intro}
Photographic text-to-face synthesis is a mainstream problem with potential applications in image  editing,  video  games,  or  for  accessibility. The task can be addressed as learning a mapping from a semantic text space describing the facial features \emph{e.g., ``Pointy Nose" and ``Waivy hair"} to the RGB pixel space. The community has traditionally addressed faces in the context of image recognition \cite{wright2009robust} where the task is to recognize the human faces from the visual descriptions of the images. Such tasks involved extracting fine-grain details, map them to a latent space and learn their distribution in the latent space.\renewcommand*{\thefootnote}{\fnsymbol{footnote}}\footnotetext{\textsuperscript{+}Equal Contribution}
\renewcommand*{\thefootnote}{\arabic{footnote}}
\setcounter{footnote}{0}

Recent advances in generative modelling \cite{goodfellow2014generative} spurred a lot of interest in the research community to generate faces by learning a mapping to the pixel space from a latent noise space. While works like BeautyGAN \cite{li2018beautygan} demonstrating style transfer on faces and face captioning using GANs \cite{Omid2018facecap} have been done but the problem of face synthesis from textual descriptions remain largely unaddressed due to following obstacles.

\begin{figure}[!t]
    \begin{tabular}{p{2cm}p{5cm}}
         \includegraphics[scale=0.7,valign=t]{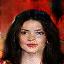} &
         The woman has high cheekbones. She has straight hair which is black in colour. She has big lips with arched eyebrows. The smiling, young woman has rosy cheeks and heavy makeup. She is wearing lipstick.
    \end{tabular}
    \caption{Example of image generated from caption in the ``zero-shot" setting.}
    \label{fig:my_label}
\end{figure}
\begin{enumerate}
    \item Widely used datasets  such as Flickr8K \cite{hodosh2013framing}, Flickr30K \cite{young2014image}, VLT2K \cite{elliott2013image},  and MS COCO \cite{lin2014microsoft} contain textual descriptions at concrete conceptual level describing broadly the object and the context without saying anything about the inferences that could be drawn from the images. While helpful these captions do not contain physical description of faces such as skin color, eyes, hairstyle, \emph{etc.} that are necessary for generating faces.
    
    \item The existing face datasets such as LFW \cite{Huang2012a} and MegaFace \cite{kemelmacher2016megaface} lack any additional description while others such as LFWA \cite{Huang2012a} and CelebA \cite{liu2015faceattributes} have a list of attributes associated with the images. Despite providing fine-grain information about faces such as \emph{``Blond Hair"} and \emph{``Arched eyebrows"} attributes requires knowledge of the domain. As a result attributes cannot be used for general purpose user end applications.
    
    \item The conditional distribution of the face (conditioned on text) is highly multimodal due to the multiple possible pixel orientations being semantically consistent with the facial features present in the text.  Presence of more fine-grained details in the facial description than scene descriptions makes learning the joint representation difficult in the ``zero-shot" setting. 
    
\end{enumerate}
In this paper, we address the aforementioned problem as learning the joint distribution of images in the pixel space and text mapped to a latent encoding space. Natural language provides a generic interface to represent information on facial features. Hence captions with information on the faces provide a way to combine the discriminative abilities of the attributes as well as the generality of natural language. We create the captions for the CelebA \cite{liu2015faceattributes} dataset from the attributes provided as the solution to dataset unavailability. We divided the captions into six sentences with each sentence capturing the features specific to certain parts of face \emph{e.g.} the first sentence captures the face outline such as high cheekbones and while the second sentence captures the hairstyle such as waivy hair (see Table~1). The automatic generation ensured the captions are free from the bias due to the subjective nature of human generated captions. The generated captions are encoded using the Skip-Thought \cite{kiros2015skip} model to better capture the facial features as well as their spatial orientation so as to maintain consistency with the general semantics of a face (``mouth should be above the nose").

The advent of GANs marked a major breakthrough in generative modelling and has become the mainstream solution to the problem of learning conditional multi-modality.  We solve the problem of learning the joint distribution of text and images by using the generator to generate the face while conditioning both generator and discriminator on the encoded facial descriptions. Apart from leveraging the property of discriminator network acting as an adaptive loss function, we explicitly provide the discriminator the sources of error as discriminator has to
differentiate whether the joint $\langle$image, text$\rangle$ pair is real or fake as mentioned in the GAN-CLS \cite{reed2016generative} algorithm. In the midst of experimentation we faced the problem of faster convergence of the discriminator loss towards 0 and to tackle the same we introduced noise in the discriminator by swapping the  real and the fake images after every three iterations. Figure 1 shows the image generated by our model for the given caption.

We evaluate our GAN model using the widely used inception score which requires around 50K generated samples. The generated samples are classified by the InceptionV3 \cite{SzegedyVISW15} model and the predicted classes are used to calculate the marginal distribution $p(y)$ and conditional distribution $p(\textbf{y}|\textbf{x})$ for all images x and classes y (see Equation 1).

\begin{equation}
  p(y) = \int_{\textbf{x}}{}p(\textbf{y}|\textbf{x})dx    
\end{equation}

Popular datasets for image synthesis from text such as Oxford-102 Flowers \cite{nilsback2008automated} and Caltech-USD Birds \cite{welinder2010caltech} contain classes with high intraclass similarity and very low interclass similarity. This property ensures that if the captions selected to generate the images (while evaluation) are uniform across classes then the inception score would reflect the clarity and diversity of the images.
We finally shows why the widely used inception score is not a good metric to evaluate the performance of GANs on the face datasets (see Section~\ref{Evaluation}).

The main contributions of this paper are as follows:
\begin{enumerate}
    \item Caption creation\footnote{We will release the captions to public for research purpose.} for CelebA dataset to facilitate face generation from textual descriptions.
    \item GAN model to synthesize faces from description of fine-grained facial features.
    \item GAN model evaluation using inception score and justification as to why it is not a good metric for face datasets.
\end{enumerate}

The rest of the paper has been organised as follows.
 Section \ref{relatedwork} discusses previous works on text to image conversion and style transfer on faces using GANs \cite{goodfellow2014generative}. Section \ref{background} provides necessary background for GANs and inception score to understand the impact of the randomness of image generation on the inception score.  Our methodology to automatically generate captions from attributes list and our network architecture of GAN \cite{goodfellow2014generative} is discussed in Section \ref{methodology}. Section \ref{Evaluation} presents the evaluation model used inferences from the inception score. This section discusses how inception score is affected by the randomness in Generated images. Finally, Section \ref{conclusion} concludes the paper and presents certain extensions of this work.

\section{Related work}\label{relatedwork}

Deep learning has led to substantial progress in the field of generative image modelling with the introduction of deep generative models such as GANs \cite{goodfellow2014generative, radford2015unsupervised}, Variational Auto Encoders \cite{kingma2013auto}, and others.

Multimodal deep learning has shown to learn relating features across modalities like text \cite{SHAH2016102}, audio \cite{Yu:2019:DCC:3309717.3281746}, visual \cite{8432497,Shah:2014:APV:2647868.2654919} and more \cite{shah2017multimodal}. One natural extension of image generation is text to image synthesis, which requires prediction of data in one modality (image) conditioned on data in another modality (text). Reed \emph{et al.}~\cite{reed2016generative} tackled this problem by using a deep convolutional generative adversarial network (DC-GAN) \cite{radford2015unsupervised} conditioned on text features encoded by a hybrid character-level convolutional recurrent neural network. Using their model they were able to produce $64\times64$ images. Zhang \emph{et al.}~ \cite{zhang2017stackgan} proposed a two stage training strategy to produce $256\times256$ images. Recently, Zhang \emph{et al.}~\cite{zhang2018photographic} proposed a GAN architecture with hierarchically-nested discriminators. This allows the authors to create $512\times512$ images. These models are usually evaluated on Oxford-102 Flowers \cite{nilsback2008automated}, Caltech-UCSD Birds \cite{welinder2010caltech} and MS-COCO datasets  \cite{chen2015microsoft}. 

Due to absence of objective function and high cost of human evaluation, text to image synthesis models are evaluated using an automated method such as Inception Score \cite{salimans2016improved}. Inception score measures both the objectiveness and diversity of generated images. It requires fine-tuning of Inception model \cite{SzegedyVISW15} pre-trained on ImageNet.

A very similar, yet relatively less researched problem is Text to Face generation which requires generation of images consisting of faces from input text description. This problem is difficult to solve mainly due to the absence of paired text and face image dataset. The current dataset Face2Text \cite{gatt2018face2text} consists of only 400 facial images and textual captions for each of them. However complex models cannot be used for such a small dataset as the generator can easily learn the entire dataset and hence will not be able to produce any results for unseen text descriptions (zero-shot setting). Though in the work \cite{T2F}, authors used a hybrid model of stackGAN and proGAN to generate faces from captions. However, the results that they have received are very poor (see Figure 2).

\begin{figure}[h]
    \centering
    \includegraphics[scale=0.4]{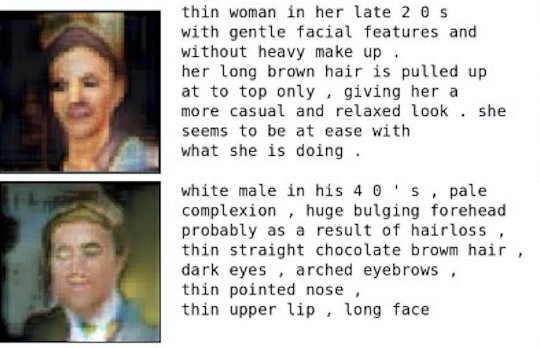}
    \caption{Existing Experiments \cite{T2F} on Face2Text dataset}
    \label{fig:my_label}
\end{figure}

To solve this problem we leverage the CelebA dataset \cite{yang2015facial} by introducing captions. These captions are generated by segregating the attributes of the images into six sentences based on structure of the face, facial hair, hairstyle, fine grain face details, accessories worn and attributes that enhance the appearance. The final caption thus created is the concatenation of all the five sentences created. However, since most images have a subset of the given attributes creating all six sentences are not always possible. Hence the length of each caption can vary widely. Due to the instability of GAN training and inconsistency of caption length, the problem of text to face synthesis becomes more difficult. We employ a variety of methods \cite{chintala2016train} such as maximizing $log(D)$ instead of minimizing $log(1-D)$, adding noise to labels for discriminator, and others to deal with the instability of GAN training. To deal with inconsistent caption length we use Skip-thought vectors \cite{kiros2015skip}. 

Face synthesis has also been done based on audio input. In WAV2PIX \cite{duartewav2pix} the authors generated face from raw audio input. They trained their model in a self-supervised approach by exploiting the audio and visual signals naturally aligned in videos. They used high quality YouTube videos for this where the speaker was expressive in both speech and signals. Also recently Karras \emph{et al.}~\cite{karras2019style} proposed GAN architecture that enables unsupervised separation of high-level attributes(e.g., pose and identity when trained on human faces) and stochastic variation in the generated images (e.g., freckles,hair). 

Building on the ideas of the previous models we provide state of the art results for the problem of text to face generation. We then provide inception score for our model by fine-tuning Inception model on CelebA dataset.

\section{Background}\label{background}
In this section we provide previous works our model builds on. We first describe how Generative Adversarial Networks (GANs) work. Then we describe GAN-CLS architecture where GANs have been used for the problem of Text to Image synthesis. We further describe Skip-Thought vectors and how they are useful for our problem. Finally we describe Inception Score and how it is evaluated.

\subsection{Generative Adversarial Networks}

Generative Adversarial Networks (GAN) is a framework that allows us to learn a function or program that can generate samples that are very similar to samples drawn from a given training distribution. It consists of a generator $G$ and a discriminator $D$ that compete in a minimax game \cite{goodfellow2014generative}. $D$ tries to distinguish between real training data and synthetic data, while $G$ tries to fool $D$. This minimax game is given by Equations 2 and 3.

\begin{equation}\label{eq:3}
    J^{(D)} = -\frac{1}{2} \mathbb{E}_{\textbf{x} \stackrel{}{\sim} p_{data}} \log D(\textbf{x}) - \frac{1}{2} \mathbb{E}_{\textbf{z}} \log(1-D(G(\textbf{z})))
\end{equation}

\begin{equation}
    J^{(G)} = -J^{(D)}
\end{equation}
where $J^{(D)}$ is the discriminator cost and $J^{(G)}$ is the generator cost, $p_{data}$ is the probability distribution of given data.
Goodfellow \emph{et al.}~\cite{goodfellow2014generative} proved that the Nash Equilibrium of this game is when samples produced by G is indistinguishable from samples coming from training data (provided G and D have enough capacity).

\subsection{Matching-aware Discriminator (GAN-CLS)}

Text-to-image synthesis can easily be modelled using conditional GANs by treating the text, image pairs as joint observations. The discriminator now has to judge the pairs as real or fake. 
In a vanilla conditional GAN, the discriminator must discriminate between real images with matching text, and synthetic images with arbitrary text. Therefore, it must implicitly learn to distinguish synthetic images and realistic images with incorrect captions.

To tackle this problem, Reed \emph{et al.}~\cite{reed2016generative} modified the discriminator by adding a third input consisting of real images with mismatched text (see Equations 4, 5, and 6).

\begin{equation}
    J^{(D)} = 
    J^{(D)}_{adv}\mathbb{E}_{\textbf{x} \stackrel{}{\sim} p_{data}} \log D(\textbf{x}, \varphi(\,\textbf{\textit{\^ t}}\,))
\end{equation}

\begin{equation}
    J^{(D)}_{adv} = 
    - \mathbb{E}_{\textbf{x} \stackrel{}{\sim} p_{data}} \log D(\textbf{x},\varphi(\,\textbf{\textit{t}}\,)) - \mathbb{E}_{\textbf{z}} \log(1-D(G(\textbf{z}, \varphi(\,\textbf{\textit{t}}\,))))
\end{equation}

\begin{equation}
    J^{(G)} = -J^{(D)}
\end{equation}
where:
\begin{description}
\item $J^{(D)}$ is the discriminator cost
\item $J^{(G)}$ is the generator cost
\item $p_{data}$ is the probability distribution of given data
\item $\varphi(\,\textbf{\textit{t}}\,)$ is the text embedding corresponding to a given image
\item $\varphi(\,\textbf{\textit{\^ t}}\,)$ is the text embedding corresponding to a different image
\end{description}

\subsection{Skip-thought Vectors}

We need to encode the given input text to learn the mapping between the text and face image. We use Skip-Thought vectors \cite{kiros2015skip} to encode the input text to a 4800 dimension vector by using the pretrained model provided by the authors. The skip-thought vectors are generated by training an encoder-decoder model. An encoder maps the sentence to a vector, whereas the decoder generates surrounding sentences from the vector. Kiros \emph{et al.} used an RNN encoder with GRU activations and an RNN decoder with Conditional GRU. These vectors obtain very good results for image retrieval task (retrieve images that are good fit to given query sentence) of MS COCO. 

\subsection{Inception Score}
The lack of objective function makes it difficult to evaluate and compare Generative Adversarial Networks. Primarily used method for evaluation is human annotation of the generated images. However based on the motivation of annotator and task setup such human evaluation can be subjective. To overcome this Salimans \emph{et al.}~\cite{salimans2016improved} proposed the inception score metric to automatically evaluate performance of GANs \cite{goodfellow2014generative}. It uses the Inception model \cite{SzegedyVISW15} to calculate the conditional distribution $p(y|x)$, and the marginal distribution $p(y)$ as show in Equation \ref{eq:7}.
\begin{equation}\label{eq:7}
    p(y) = \int_{\textbf{x}}{}p(\textbf{y}|\textbf{x})dx
\end{equation}
The final inception score in calculated as the KL divergence of these distributions (see Equation 8).
\begin{equation}
    \mathbb{E}_{\textbf{x}}\textbf{KL}(p(y|\textbf{x})||p(y))
\end{equation}
where $\textbf{x}$ is random variable for image and $\textbf{y}$ for classes. The conditional distribution $p(\textbf{y}|\textbf{x})$ captures the clarity of the generated images. The marginal distribution  $p(\textbf{y})$ captures the diversity of the GAN model. A higher inception score corresponds to a skewed $p(\textbf{y}|\textbf{x})$ as the inception model \cite{SzegedyVISW15} predicts the class for the given image with high confidence. Moreover the marginal distribution $p(\textbf{y})$ should be uniform reflecting that the GAN model is not biased towards any particular class. For a good model, $p(\textbf{y}|\textbf{x})$ should have high entropy while $p(\textbf{y})$ should
have low entropy.
\begin{figure*}[h]
\begin{center}
  \includegraphics[scale=0.45]{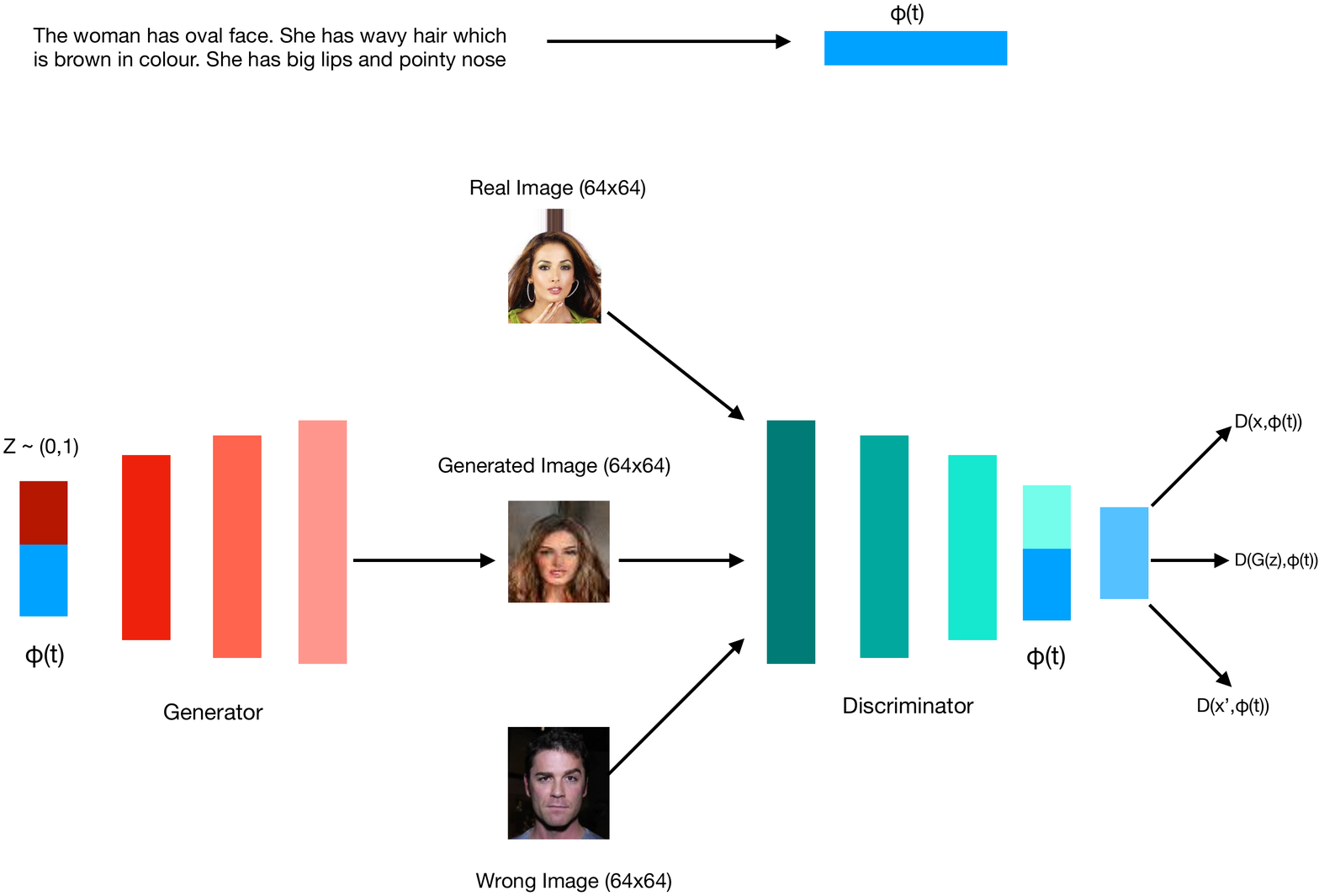}
  \caption{Our text conditional-convolutional GAN architecture conditioned on captions. The real and fake images are swapped after every third iteration.}
  \label{fig:arch}
\end{center}
\end{figure*}

\section{Methodology}\label{methodology}
In this section, we describe our algorithm for automatic caption generation along with our modeling of the problem of text-to-face as learning conditional distribution of faces (conditioned on text).
We begin by providing the algorithm and justification as to why our algorithm captures all the features of an image in meaningful and versatile captions. 
We then explain why the problem of mapping text to faces is unsupervised learning of conditional representation and how conditional multimodality comes into picture. Finally we show how to model it using GANs \cite{goodfellow2014generative} and our modifications to prevent the faster convergence of discriminator. Figure 3 shows the architecture of  our Text Conditional-Convolutional GAN which is conditioned on captions. 

\subsection{Caption Generation}
To convert the attribute list provided for the images in the CelebA \cite{liu2015faceattributes} dataset to meaningful captions, we create six group of features in response to six questions which progressively describe the face starting from the face outline to the facial features which enhance the appearance (see Table \ref{table1}). Apart from these set of attributes, we use words describing the gender of the celebrity, \emph{e.g., ``she", ``he", and other.}

\begin{table}[!h] 
\caption{Questions and the corresponding set of attributes as response }
\begin{tabular}{|p{3.5cm}|p{4.2cm}|}
    \hline 
    \textbf{Questions for Facial Groups} & \textbf{Facial Attributes used for Answers}\\ \hline
    What is the structure of the face? & Chubby face, Double Chin, Oval face, High cheekbones \\[0.1cm] \hline
    
    What is the facial hairstyle does the person sport? & 5 O Clock Shadow, Goatee, Mustache, Sideburns \\[0.1cm] \hline
    
    What hairstyle does the person sport? & Bald, Straight hair, Black hair, Blond hair, Brown hair, Gray hair, Bangs, Wavy hair, Receding hairline. \\[0.1cm] \hline
    
    What is the description of the other facial features? & Big lips, Big nose, Pointy nose, Narrow eyes, Arched eyebrows, Bushy eyebrows, Mouth slightly open. \\[0.1cm] \hline
    
    What are the attributes that enhance the appearance? & Young, Attractive, Smiling, Pale skin, Rosy cheeks, Heavy makeup. \\[0.1cm] \hline
    
    What are the accessories worn? & Earrings, Hat, Necklace, Necktie, Eyeglasses, Lipstick \\[0.1cm] \hline
\end{tabular}
\label{table1}
\end{table}

The questions are so aligned to assist the Generator in GANs \cite{goodfellow2014generative} to build the face by first learning to create the face outline, then add hair in the specified hairstyle followed by creating eyes, nose \emph{etc.,} then enhance appearance with the features like ``young", ``attractive" and finally add the specified accessories in the captions.

We maintain a dictionary with attributes as the keys with corresponding values being the  set of words to replace them in the sentence, \emph{e.g.,} ``\textit{Mouth\_Slightly\_Open}"$\colon$``\textit{slightly open mouth}".
In order to create a sentence from a given set of attributes we create a queue. We first add the start of the sentence to the queue (\emph{e.g.,} ``He sports a"). Then we add the corresponding values for the first feature to the queue (\emph{e.g.,} 5 o'clock shadow). For every subsequent attributes we add a conjunction or punctuation to the queue before the attribute, provided there is already an attribute at the end of the queue. 
Otherwise we add the next attribute directly (see Algorithm 1). 
Suppose the list of attributes has ``goatee" and ``mustache" as the features describing facial hair. The queue initially contains ``He sports a" (notice that the back of queue has ``a" which is not an attribute). We add the first feature i.e goatee directly. Queue now is ``He sports a goatee". Next feature is mustache. Since the back of queue has an attribute therefore we add a conjunction (\emph{i.e.,} ``and") to the queue before adding mustache. So the final queue is ``He sports a goatee and mustache".
Our algorithm has O(\,$nl$\,) running time complexity, where $n$ is the number of images and $l$ is the length of the attributes list. For CelebA dataset \cite{liu2015faceattributes}, $l=40$ hence the running time becomes O(\textbf{$n$}) which is linear in \textbf{$n$}.

\begin{algorithm}[!h] \label{alg:1}
\caption{Caption Creation For Facial Hair attributes}\label{alg:captn}
\begin{algorithmic}[1]

\Procedure{Facial Hair Caption}{$isPresent$}

\State $Q \gets $\{He, sports, a\} \Comment{Q is a queue}

\State $L \gets $ \{5 o\'clock shadow, Goatee, Mustache, Sideburns\}
\State $conjunction[Goatee] \gets$ `,' 
\State $conjunction[Mustache] \gets and $
\State $conjunction[Sideburns] \gets with $

\ForAll{$l \in L$}
    \If{$isPresent(l)$}
        \If{$Q.back() = a$}
            \If{$l \neq sideburns$}
                \State $Q.push(l)$
            \Else
                \State $Q.clear()$
                \State $Q \gets $\{He has sideburns\}
            \EndIf
        \Else
            \State $Q.push(conjunction[l])$
            \State $Q.push(l)$
        \EndIf
    \EndIf
\EndFor
    
\State \textbf{return} $Q$
\EndProcedure
\end{algorithmic}
\end{algorithm}

\subsection{Network Architecture}
The generator network is represented as ${\textit{G}:{R^Z}\times{R^T}\rightarrow{}R^I}$ and the discriminator as $\textit{D}:{R^I}\times{R^T}\rightarrow{}(0,1)$ where the $Z$ is the dimension of the noise vector input to the generator, $T$ is the dimension of the skip-thought embedding of the caption and $I$ is the dimension of the generated image. We sample the input noise $\mathscr{Z}$ $\in{R^Z}\sim{U(0,1)}$ of dimension 100 and then encode the text caption \textit{t} using skip-thought encoder $\varphi(\textit{t})$ (we used 4800 as the dimension of encoding).
We reduced the dimension of the text encoding $\varphi(\textit{t})$ to  256 using fully connected layers followed by leaky RELU activation. We then concatenate the reduced encoding $\varphi(\textit{t})$ to the noise $\mathscr{Z}$ to form a vector $\theta$ of length 356 as an input to the generator.

The generator is a deconvolutional network with a projection operations, 4 deconvolutional layers and finally a tanh layer. Convolutional layers are followed by batch normalization and leaky RELU activation. The generator first projects $\theta$ to a vector $\theta_{proj}$ of dimension 8192 (see Equation \ref{eq:9}).

\begin{equation}\label{eq:9}
    \theta_{proj} = W^T\theta+B
\end{equation}
where $W$ is the projection matrix of dimension $356\times8192$ and $B$ is the bias. $\theta_{proj}$ is then reshaped into a tensor of dimenison height 4, width 4 and 512 channels. Further the deconvolutinoal layers decrease the number of channels by a factor of 2 and increase the height and width by the same. The last deconvolutional layer converts the tensor output of the fourth layer (with height 32, width 32 and 64 channels) followed by tanh to a $64\times64\times3$ RGB image.

The discriminator is a convolutional network with four convolutional layers having strides of 2, dimension expansion (after 4th convolutional layer) and finally a sigmoid layer. Convolutional layers are followed by batch normalization and leaky RELU activation. The first convolutional layer converts a RGB image of dimension $64\times64\times3$ to a tensor of height 32, width 32 and 64 channels. The next three convolutional layers progressively decrease the height and width by a factor of 2 and increase the number of channels by a factor of 2. The resulting tensor $\gamma$ is of dimension $4\times4\times512$. 
Then the dimension of $\varphi(\textit{t})$ is expanded to $4\times4\times256$
and concatenated to $\gamma$ along third dimension which is convolved over over by the final convolutional layer. The output of final convolutional layer is passed to sigmoid layer to generate a confidence score between (0,1).

GANs \cite{goodfellow2014generative} experience the problem of faster convergence of the discriminator over generator leading to no learning of generator. For conditional GANs, this becomes even more difficult as the generator has to generate images in the pixel space while maintaining semantic similarity in the text space. When the discriminator learns faster than generator $D(x)\approx1$ and $D(G(\varphi(\,\textbf{\textit{t}})))\approx0$. Equations \ref{eq:10} and \ref{eq:11} show how the $\log$ losses converge to 0.
\begin{equation}\label{eq:10}
    \log(D(x))\approx0
\end{equation}
\begin{equation}\label{eq:11}
    \log(1-D(G(\varphi(\,\textbf{\textit{t}}))))\approx0
\end{equation}
Hence in Equation \ref{eq:3}, $J^{(D)}\approx0$ and the generator cannot learn anything from thereon.
To tackle this, we swapped the real and the generated images for the discriminator after every three iterations. This fools the discriminator into believing that generated images are real, slowing down the learning and providing essential time for generator to catch up to discriminator.

\begin{figure*}
    \centering
    \begin{tabular}{p{2cm}p{5cm}@{\hskip 1.5cm}p{2cm}p{5cm}}
        \includegraphics[scale = 0.7,valign=t]{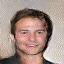} & 
        The man has oval face and high cheekbones. He has wavy hair which is brown in colour. He has a slightly open mouth.
        The young attractive man is smiling. & 
        \includegraphics[scale=0.7,valign=t]{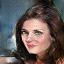} &
        The woman has high cheekbones. She has wavy hair. The young attractive woman has heavy makeup. She's wearing a necklace and lipstick. \\[1cm]
        
        \includegraphics[scale=0.7,valign=t]{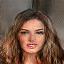} &
        The woman has oval face. She has wavy hair which is brown in colour. She has big lips and pointy nose with arched eyebrows and a slightly open mouth. The young attractive woman has heavy makeup. She's wearing lipstick. &
        \includegraphics[scale=0.7,valign=t]{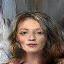} &
        The woman has high cheekbones. She has wavy hair. She has arched eyebrows. The young attractive woman has heavy makeup. She's wearing lipstick. \\[1.5cm]
        
        \includegraphics[scale=0.7,valign=t]{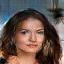} &
        The woman has oval face. She has straight hair which is brown in colour. The smiling, young attractive woman has heavy makeup. She's wearing lipstick. &
        \includegraphics[scale=0.7,valign=t]{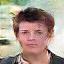} &
        The man's hair is brown in colour. The man looks young. \\[2cm]
        
        \includegraphics[scale=0.7,valign=t]{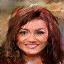} &
        The woman has oval face and high cheekbones.She has straight hair which is brown in colour. She has big lips and narrow eyes with arched eyebrows and a slightly open mouth. The smiling, young attractive woman has heavy makeup. She's wearing lipstick. &
        \includegraphics[scale=0.7,valign=t]{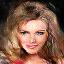} &
        The woman has wavy hair which is blond in colour. She has big lips with arched eyebrows and a slightly open mouth. The young attractive woman has rosy cheeks and heavy makeup. She's wearing lipstick.\\[1.2cm]
        
        \includegraphics[scale=0.7,valign=t]{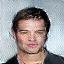} &
        The man sports a 5 o'clock shadow. His hair is black in colour. He has big nose with bushy and arched eyebrows. The man looks attractive. &
        \includegraphics[scale=0.7,valign=t]{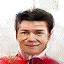} &
        The man sports a 5 o'clock shadow and mustache. He has a receding hairline. He has big lips and big nose, narrow eyes and a slightly open mouth. The young attractive man is smiling. He's wearing necktie.\\[1cm]
        
        \includegraphics[scale=0.7,valign=t]{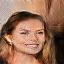} &
        The woman has oval face and high cheekbones. Her straight hair has shades of blond.She has a slightly open mouth. The smiling, young attractive woman has heavy makeup. She's wearing lipstick. &
        \includegraphics[scale=0.7,valign=t]{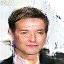} &
        The man has straight hair.He has arched eyebrows.The man looks young and attractive.He's wearing necktie. \\[1.5cm]
        
        \includegraphics[scale=0.7,valign=t]{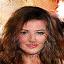} &
        The woman has high cheekbones. She has wavy hair which is brown in colour. She has big lips with arched eyebrows. The smiling, young woman has rosy cheeks and heavy makeup. She is wearing lipstick. &
        \includegraphics[scale=0.7,valign=t]{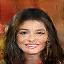} &
        The woman has high cheekbones. She has straight hair which is brown in colour. She has arched eyebrows and a slightly open mouth. The smiling, young attractive woman has heavy makeup. She is wearing lipstick.
        
    \end{tabular}
    \caption{Qualitative results for visual inspection. Above images contain selected few features and are generated in the ``zero-shot" setting i.e. unseen text.}
    \label{fig:4}
\end{figure*}

\section{Evaluation and Results}\label{Evaluation}
We ran our model on the 10000 random selected images from CelebA \cite{liu2015faceattributes} dataset with our created captions for 200 epochs. The training set consists of 7500 images and the testing set consist of 2500 images. We used batches of the dataset to train the model with a batch size of 64. Learning rate for generator was set to 0.0002 and for discriminator was 0.0001. We used Adam \cite{Kingma2015adam} with $\beta_1=0.5$ and $\beta_2=0.5$ for both generator and discriminator. We used the Inception score \cite{salimans2016improved} to evaluate the performance of our model and also present the generated images for visual inspection (see Figure \ref{fig:4}). The identities of the celebrities were used as the classes. We kept the number of captions from every class uniform to ensure that the generated images are not biased towards a specific class. Non-uniform distribution of captions over classes could lead to generation of more images belonging to the class with higher captions which makes class distribution (conditioned on generated images) skewed giving a poor inception score. Such results could not lead to any conclusion as the same model could lead to uniform class distribution (conditioned on generated images) giving a good inception score. 

\subsection{Results and Inferences}
Our model gave an inception score of \textbf{1.4}$\pm$\textbf{0.7} over 5 iterations of evaluation. The images generated from our model show promising results. Our model is not facing mode collapse which can be observed in the last two images of Figure \ref{fig:4} which are significantly different even though they have very similar captions. The high variance \textbf{0.7} suggests randomness in the marginal distribution as computed by Equation \ref{eq:12}.
\begin{equation}\label{eq:12}
    p(\textbf{y}) = \int_{x}{}p(\textbf{y}|\textbf{x}=G(\textbf{z}))dz
\end{equation} 
In some iterations, predicted classes are uniformly distributed while for others they are highly skewed.
The low inception score shows that the marginal distribution $p(\textbf{y})$ has high entropy and very similar to $p(\textbf{y}|\textbf{x})$ $\forall$ image $\textbf{x}$, class $\textbf{y}$ and text encoding $\textbf{z}$.

Popular datasets such as Oxford-102 Flowers \cite{nilsback2008automated} and Caltech-USD Birds \cite{welinder2010caltech} have classes such that the captions for images have high intraclass similarity and very low interclass similarity. Descriptions of ``Lily" \emph{e.g. ``This flower is white and pink in color, with petals that have veins"} shows clear semantic dissimilarity with that of ``Sunflower" \emph{e.g. ``The flower has yellow petals and the center of it is brown"}.  For these datasets while calculating inception score, if the captions are uniformly distributed over classes and the model is good then the generated images would be classified with high confidence with uniform class distribution. Han Zangh \emph{et al.} \cite{zhang2017stackgan} calculated an inception score of 2.88$\pm$0.04 for Oxford-102 Flowers \cite{nilsback2008automated} and 2.66$\pm$0.06 for Caltech-USD birds \cite{welinder2010caltech}.

\begin{figure}[!h]
    \centering
   \begin{tabular}{p{2cm}p{5cm}}
        \includegraphics[scale=0.3,valign=t]{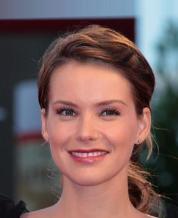} &  The woman has oval face and high cheekbones. She has straight hair which is brown in colour. She has arched eyebrows and a slightly open mouth. The smiling, young attractive woman has heavy makeup. She is wearing earrings and lipstick. \\[0.3cm]
        
        \includegraphics[scale=0.3,valign=t]{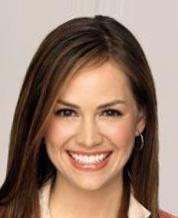} & The woman has oval face and high cheekbones. She has straight hair which is brown in colour. She has big lips and narrow eyes with arched eyebrows and a slightly open mouth. The smiling, young attractive woman has heavy makeup. She is wearing earrings and lipstick.
   \end{tabular}
   \caption{Similarity in the facial features for celebs with different identities.}
   \label{fig:3}
\end{figure}

Person's identity or any other class based on attributes is a very poor choice for classifying the images as the captions have high interclass similarity (due to high possibility of similar facial features being present across classes) as shown in Figure \ref{fig:3}. For instance, in this figure both captions are almost similar but they belong to two different celebrities. As a result when conditioned on caption \textbf{\textit{t}} the model could randomly generate semantically similar face $G(\varphi(\,\textbf{\textit{t}}\,))$ belonging to any of the classes (having captions capturing similar facial features as the query caption). This randomness could result in generation of a lot of images for a few classes while very few for others. As discussed above, even a good inception score in some iteration of the experiment cannot be used to infer better performance of GANs \cite{goodfellow2014generative} in terms of producing quality images semantically similar with query captions. This argument is strengthened by the fact that the generated images are very good and semantically similar to the textual descriptions.

\section{Conclusion and Future work}\label{conclusion}
In this work we presented captions for the CelebA dataset to facilitate face synthesis from text. We then used Generative Adversarial Network to learn the conditional multimodality in synthesis of face from captions. Finally we demonstrated why inception score used to measure the performance of GANs \cite{goodfellow2014generative} fails to evaluate their performance on our dataset.

We plan on extending the work in the following directions:
\begin{enumerate} 
    \item Improve the selection of the wrong image for the GAN-CLS \cite{reed2016generative} algorithm. Currently, we randomly select images from the dataset as wrong image. One possibility is to select the wrong caption for real image rather than selecting the wrong image. This could be done by selecting the caption having the lowest cosine similarity with the caption of the real image.
    
    \item Explore better language models such as BERT, analyze and compare performance of other GAN architectures with our model for face generation from captions.
    
    \item Propose a better evaluation metric to capture the semantic similarity of the generated faces with their captions, without using the classes.
    
    \item Improving the resolution of the generated faces e.g. $128\times128$ and $256\times256$ faces.
\end{enumerate}

\section{Acknowledgement}\label{acknowledgement}
Rajiv Ratn Shah is partly supported by the Infosys Center of AI, IIIT Delhi and ECRA Grant by SERB, Govt. of India. This work was partially supported by JSPS Grant-in-Aid for Scientific Research (C) under Grant No.  1 9 K 1 1 9 8 7.

\bibliographystyle{IEEEbib}
\bibliography{bigmm2019template}

\end{document}